\documentclass[runningheads]{llncs}
\usepackage{graphicx}
\usepackage{todonotes}
\usepackage{hyperref}
\usepackage{url}
\usepackage{cite}
\usepackage{bbm}
\usepackage[perpage]{footmisc}
\usepackage{amsmath}
\usepackage{amssymb}
\usepackage{multirow}
\usepackage{algorithm,algorithmic}
\usepackage{comment}
\usepackage{graphicx}
\usepackage{bbm}
\usepackage{bm}
\usepackage{dsfont}
\usepackage{mathtools}
\usepackage{adjustbox}

\begin{document}
\title{Semi-supervised Meta-learning with Disentanglement for Domain-generalised Medical Image Segmentation}
\titlerunning{Semi-supervised Meta-learning for Domain Generalisation}

\author{Xiao Liu\inst{1}, Spyridon Thermos\inst{1}, Alison O'Neil\inst{1,3} \and
Sotirios A. Tsaftaris\inst{1,2}} 
\authorrunning{X. Liu et al.}

\institute{School of Engineering, University of Edinburgh, Edinburgh EH9 3FB, UK \and
The Alan Turing Institute, London, UK \and
Canon Medical Research Europe Ltd., Edinburgh, UK\\
\email{Xiao.Liu@ed.ac.uk, SThermos@ed.ac.uk, Alison.ONeil@mre.medical.canon, S.Tsaftaris@ed.ac.uk}}
\maketitle              
\sloppy
\begin{abstract}
Generalising deep models to new data from new centres (termed here domains) remains a challenge. This is largely attributed to shifts in data statistics (domain shifts) between source and unseen domains. Recently, gradient-based meta-learning approaches where the training data are split into meta-train and meta-test sets to simulate and handle the domain shifts during training have shown improved generalisation performance. However, the current fully supervised meta-learning approaches are not scalable for medical image segmentation, where large effort is required to create pixel-wise annotations. Meanwhile, in a low data regime, the simulated domain shifts may not approximate the true domain shifts well across source and unseen domains. To address this problem, we propose a novel semi-supervised meta-learning framework with disentanglement. We explicitly model the representations related to domain shifts. Disentangling the representations and combining them to reconstruct the input image allows unlabeled data to be used to better approximate the true domain shifts for meta-learning. Hence, the model can achieve better generalisation performance, especially when there is a limited amount of labeled data. Experiments show that the proposed method is robust on different segmentation tasks and achieves state-of-the-art generalisation performance on two public benchmarks. Code is publicly available at: \url{https://github.com/vios-s/DGNet}.

\end{abstract}

\section{Introduction}
Despite recent progress in medical image segmentation  \cite{bernard2018deep, isensee2017automatic, chen2020frontiers}, inference performance on unseen datasets, acquired from distinct scanners or clinical centres, is known to decrease  \cite{prados2017spinal, mnms}. Such reduction is mainly caused by shifts in data statistics between different clinical centres i.e. \textit{domain shifts} \cite{zhang2020generalising}, due to variation in patient populations, scanners, and scan acquisition settings \cite{tao2019deep}. The variation in population impacts the underlying anatomy and pathology due to factors such as gender, age, ethnicity, which may differ for patients in different locations \cite{li2020self, wang2020meta, esther2021fairness}. The variation in scanners and scan acquisition settings impacts the characteristics of the acquired image, such as brightness and contrast \cite{zhang2020generalising}.

The naive approach to handling domain shift is to acquire and label as many and diverse data as possible, the cost implications and difficulties of which are known to this community. Alternatively one can train a model on source domains to generalise for a target domain with some information on the target domain available i.e. domain adaptation \cite{bian2020uncertainty} such as cross-site MRI harmonisation \cite{pomponio2020harmonization} to enforce the source and target domains to share similar image-specific characteristics \cite{dewey2020disentangled}. A more strict alternative is to \textit{not use any} information for the target domain, known as \textit{domain generalisation} \cite{li2018learning}. Herein, we focus on this more challenging and more widely applicable approach.

In domain generalisation, the overarching goal is to identify suitable representations that encode information about the task at hand whilst being insensitive to domain-specific information. There are several active research directions aiming to address this goal, including: direct augmentation of the source domain data \cite{zhang2020generalising}, feature space regularisation \cite{muandet2013domain, li2018deep, carlucci2019domain, huang2021fsdr, zhao2020domain}, alignment of the source domain features or output distributions \cite{li2020domain}, and learning domain-invariant features  with gradient-based meta-learning \cite{li2018domain, dou2019domain, liu2020shape}. Of the above, gradient-based meta-learning methods have the advantage of not overfitting to dominant source domains which account for the more populous data in the training dataset \cite{dou2019domain}. Gradient-based meta-learning \cite{li2018learning, li2020difficulty} exploits an episodic training paradigm \cite{li2019episodic} by splitting the source domains into meta-train and meta-test sets at each iteration. The model is trained to handle domain shift by simulating it during training. By using constraints to implicitly eliminate the information related to the simulated domain shifts, the model can learn to extract domain-invariant features. Previous work introduced different constraints in a fully supervised setting e.g. global class alignment and local sample clustering \cite{dou2019domain}, shape-aware constraints \cite{liu2020shape} or simply the task objective \cite{li2018learning, khandelwal2020domain}, where \cite{khandelwal2020domain} extends \cite{li2018learning} to medical image segmentation.\footnote{With the exception of \cite{sharifi2020domain} which clusters unlabeled data to generate pseudo labels, but unfortunately is not applicable to segmentation.} These approaches do not scale in medical image segmentation as pixel-wise annotation is time-consuming, laborious, and requires expert knowledge. Meanwhile, in a low data regime where centres only provide a few labeled data samples, these methods may only learn to extract domain-invariant features from an under-represented data distribution \cite{zhang2020generalising, khandelwal2020domain}. In other words, the simulated domain shifts may not well approximate the true domain shifts between source and unseen domains.

\begin{figure}[t]
\includegraphics[width=\textwidth]{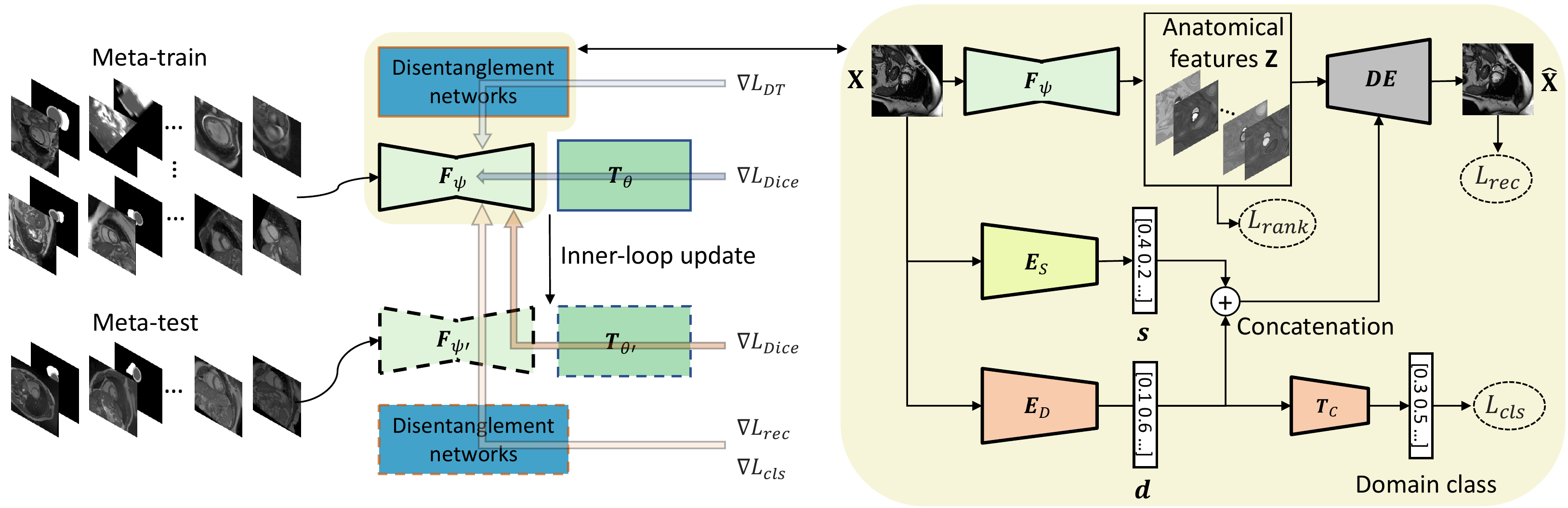}
\centering
\caption{At each iteration, the training dataset is split into meta-train and meta-test sets including labeled and unlabeled data. A feature network $\bm{F}_\psi$ extracts features $\mathbf{Z}$ for a task network $\bm{T}_\theta$ to predict segmentation masks. The model is trained in a semi-supervised setting, where $\mathcal{L}_{DT}$, $\mathcal{L}_{rec}$ and $\mathcal{L}_{cls}$ do not require pixel-wise annotation. In the inner-loop update, $\psi'$ and $\theta'$ are computed for the meta-test step (see Eq. \ref{Eqa::meta-train}). Finally, all the gradients are computed to update $\bm{F}_\psi$ and $\bm{T}_\theta$ as in Eq. \ref{Eqa::meta-test}. The disentanglement networks decompose image $\mathbf{X}$ to common $\mathbf{s}$ and specific to the domain $\mathbf{d}$ representations to be disentangled with $\mathbf{Z}$ for meta-train and meta-test sets with the constraints ($\mathcal{L}_{DT}$ and $\mathcal{L}_{rec}$ and $\mathcal{L}_{cls}$). See Section \ref{sec:method} for loss definitions.} \label{fig::model}
\end{figure}

To address this problem, we propose to explicitly disentangle the representations related to domain shifts for meta-learning as we illustrate in Fig. \ref{fig::model}. Learning these complete and sufficient representations \cite{achille2018emergence} via reconstruction brings the benefit of unsupervised learning, thus we can better simulate the domain shifts by also using unlabeled data from any of the source domains. We consider two sources of shifts: one due to scanner and scan acquisition setting variation, and one due to population variation. Because our task is segmentation, we want to be sensitive to changes in anatomy but insensitive to changes in imaging characteristics be it some common across domains or domain-specific. We use spatial (grid-like) features as a representation of anatomy ($\mathbf{Z}$) and two vectors ($\mathbf{s}$,$\mathbf{d}$) to encode common or domain-specific imaging characteristics. We apply specific design and learning biases to disentangle the above. For example, a spatial $\mathbf{Z}$ is equivariant to segmentation and this has been shown to improve performance \cite{huang2018multimodal, chartsias2019factorised}. We further encourage $\mathbf{Z}$ to be disentangled from $\mathbf{s}$ and $\mathbf{d}$ by exploiting also low-rank regularisation \cite{li2020domain}. Gradient-based meta-learning also encourages $\mathbf{Z}$, $\mathbf{s}$, and $\mathbf{d}$ to generalise well to unseen domains whilst at the same time improves (implicitly) their disentanglement. Our \textbf{contributions} are summarised as follows: \\
\noindent \textbf{1.} We propose the first, to the best of our knowledge, semi-supervised domain-generalisation framework combining meta-learning and disentanglement. \\
\noindent \textbf{2.} Use of low-rank regularisation as a learning bias to encourage better disentanglement and hence improved generalisation performance. \\
\noindent \textbf{3.} Extensive experiments on cardiac and gray matter datasets show improved performance over several baselines especially for the limited annotated data case.

\section{Proposed method}
\label{sec:method}
Consider a multi-domain training dataset $\mathcal{D}=\{\mathbf{X}_{i}^{k}, \mathbf{Y}_{i}^{k}\}_{i=1}^{N_k}, k \in \{1,2,\cdots,K\}$ that is defined on a joint space $\mathcal{X} \times \mathcal{Y}$, where $\mathbf{X}_{i}^{k}$ is the $i^{th}$ training datum from the $k^{th}$ source domain with corresponding ground-truth segmentation mask $\mathbf{Y}_{i}^{k}$, and $N_k$ denotes the number of training samples in the $k^{th}$ source domain. We aim to learn a model containing a feature network $\bm{F}_\psi: \mathcal{X} \to \mathcal{Z}$ to extract the anatomical features $\mathbf{Z}$ and a task network $\bm{T}_\theta: \mathcal{Z} \to \mathcal{Y}$ to predict the segmentation masks, where $\psi$ and $\theta$ denote the network parameters.

\subsection{Gradient-based meta-learning for domain generalisation}
In gradient-based meta-learning for domain generalisation, the domain shift is simulated by training the model on a sequence of episodes \cite{li2018domain, li2019episodic}. Specifically, the meta-train set $\mathcal{D}_{tr}$ and the meta-test set $\mathcal{D}_{te}$ are constructed by randomly splitting the source domains $\mathcal{D}$ for each iteration of training. Each iteration comprises a meta-train step followed by a meta-test step. For the meta-train step, the parameters $\psi$ and $\theta$ of $\bm{F}_\psi$ and $\bm{T}_\theta$ are calculated by optimising the meta-train loss $\mathcal{L}_{meta-train}$ with data from $\mathcal{D}_{tr}$ (inner-loop update), as defined by:
\begin{equation}
    (\psi', \theta') = (\psi, \theta) - \alpha \nabla_{\psi, \theta} \mathcal{L}_{meta-train}(\mathcal{D}_{tr};\psi, \theta),
    \label{Eqa::meta-train}
\end{equation}
where $\alpha$ is the learning rate for the meta-train update step. Typically, $\mathcal{L}_{meta-train}$ is the task objective, e.g. the Dice loss \cite{dice1945measures} for a segmentation task. This step rewards accurate predictions on the meta-train source domains. For the meta-test step, the meta-test source domains $\mathcal{D}_{te}$ are processed by the updated parameters $(\psi', \theta')$ and the model is expected to contain certain properties quantified by the $\mathcal{L}_{meta-test}$ loss. $\mathcal{L}_{meta-test}$ is computed using the updated parameters $(\psi', \theta')$, whilst the gradients are computed towards the original parameters $(\psi, \theta)$. The final objective is defined as:
\begin{equation}
    \operatorname*{argmin}_{\psi, \theta} \mathcal{L}_{meta-train}(\mathcal{D}_{tr};\psi, \theta) + \mathcal{L}_{meta-test}(\mathcal{D}_{te};\psi', \theta').
    \label{Eqa::meta-test}
\end{equation}

The intuition behind this scheme is that the model should not only perform well on the source domains, but its future updates should also generalise well to unseen domains. Below, we will describe our meta-train and meta-test objectives but first we present how we disentangle representations related to domain shifts.

\subsection{Learning disentangled representations}
To model appearance in a single-domain setting, typically a single vector-based variational representation is used \cite{chartsias2019factorised}. Here, due to our multi-domain setting, inspired by \cite{yu2019multi, ilse2020diva}, we separately encode domain-specific imaging characteristics as an additional vector-based variational representation. Hence, we aim to learn two independent vector representations, where one ($\mathbf{s}$) captures common imaging characteristics across domains and the other one ($\mathbf{d}$) captures specific imaging characteristics for each domain. In addition, we encode spatial anatomy information in a separate representation $\mathbf{Z}$, which we encourage to be disentangled from $\mathbf{s}$ and $\mathbf{d}$.

In particular, the input image $\mathbf{X}$ is first encoded in a common (appearance) representation $\mathbf{s} = \bm{E}_S(\mathbf{X})$, and a domain representation $\mathbf{d} = \bm{E}_D(\mathbf{X})$ that is followed by a shallow domain classifier $\bm{T}_C(\mathbf{d})$ which predicts the source domain ($\hat{\mathbf{c}}$) label of $\mathbf{X}$. Then, a decoder $\bm{DE}$ combines the extracted features $\mathbf{Z}=\bm{F}_\psi(\mathbf{X})$ and the representations $\mathbf{s}$ and $\mathbf{d}$ to reconstruct the input image, i.e. $\hat{\mathbf{X}} = \bm{DE}(\mathbf{Z}, \mathbf{s}, \mathbf{d})$. Note that $\bm{DE}$ combines $\mathbf{Z}$ and $\mathbf{s},\mathbf{d}$ using adaptive instance normalisation (AdaIN) layers~\cite{huang2017arbitrary}. As shown in \cite{huang2018multimodal}, AdaIN improves disentanglement and encourages $\mathbf{Z}$ to encode spatially equivariant information, i.e. anatomical information useful for segmentation, and $\mathbf{s},\mathbf{d}$ to only encode common or domain-specific appearance.

To achieve such ``triple" disentanglement we consider several losses: \textbf{1)} KL divergences $\mathcal{L}_{KL}(\mathbf{s}, N(0,1)), \mathcal{L}_{KL}(\mathbf{d}, N(0,1))$ to induce a Gaussian $N(0,1)$ prior in $\mathbf{s}$ and $\mathbf{d}$, encouraging the representations to be robust on unseen domains \cite{higgins2016beta}; \textbf{2)} Hilbert-Schmidt Independence Criterion (HSIC) loss $\mathcal{L}_{HSIC}{(\mathbf{s}, \mathbf{d})}$, to force $\mathbf{s}$ and $\mathbf{d}$ to be independent from each other~\cite{ma2020hsic}; \textbf{3)} a classification loss $\mathcal{L}_{cls}{(\mathbf{c}, \hat{\mathbf{c}})}$ such that the domain representation $\mathbf{d}$ is highly correlated with the domain-specific information \cite{ilse2020diva}; and \textbf{4)} a reconstruction loss $\mathcal{L}_{rec}{(\mathbf{X}, \hat{\mathbf{X}})}$, defined as the $\ell_1$ distance between $\mathbf{X}$ and $\hat{\mathbf{X}}$, to learn representations without supervision \cite{chartsias2019factorised, huang2018multimodal}. 

We further encourage the extracted features $\mathbf{Z}$ to be equivariant across the meta-train source domains and improve disentanglement between $\mathbf{Z}$ and $\mathbf{s}$, $\mathbf{d}$ by applying rank regularisation \cite{li2020domain}. Specifically, consider a batch $\{\mathbf{X}_{i_1}^{1}, \mathbf{X}_{i_2}^{2}, \cdots, \mathbf{X}_{i_{K_{tr}}}^{K_{tr}}\}$ from $K_{tr}$ meta-train source domains, and $K_{tr}$ features $\{\mathbf{Z}_{i_1}^{1}, \mathbf{Z}_{i_2}^{2}, \cdots, \mathbf{Z}_{i_{K_{tr}}}^{K_{tr}}\}$ extracted using the feature network $\bm{F}_\psi$. By flattening and concatenating these features, we end up with a matrix $\mathbb{Z}$ with dimensions [$C$, $K_{tr}\times H\times W$], where $C,H,W$ denote the number of channels, height, and width of $\mathbb{Z}$. Then, by forcing the rank of $\mathbb{Z}$ to be $m$ (i.e.~the number of the segmentation classes), $\mathbf{Z}$ is encouraged to encode only globally-shared information across $K_{tr}$ source domains in order to predict the segmentation mask as discussed in~\cite{li2020domain}. We achieve that by minimising the ($m+1$)$^{th}$ singular value $\sigma_{m+1}$ of $\mathbb{Z}$, i.e. $\mathcal{L}_{rank} = \sigma_{m+1}$. Overall, $\mathcal{L}_{DT}$ is defined as:
\begin{equation}
\begin{aligned}
        \mathcal{L}_{DT} = & \lambda_{rank} \mathcal{L}_{rank}(\mathbf{Z}) + \lambda_{KL} (\mathcal{L}_{KL}(\mathbf{s}, N(0,1))+\mathcal{L}_{KL}(\mathbf{d}, N(0,1))) \\ &+ \lambda_{rec} \mathcal{L}_{rec}{(\mathbf{X}, \hat{\mathbf{X}})} + \lambda_{HSIC} \mathcal{L}_{HSIC}{(\mathbf{s}, \mathbf{d})} + \lambda_{cls} \mathcal{L}_{cls}{(\mathbf{c}, \hat{\mathbf{c}})},
\end{aligned}
\end{equation}
where $\mathbf{c}$ is the domain label. We adopt hyperparameter values according to our extensive early experiments and discussion from \cite{chartsias2019factorised, li2020domain} as $\lambda_{rank}=0.1$, $\lambda_{KL}=0.1$, $\lambda_{rec}=1$ and $\lambda_{cls}=1$. Note that all the losses do not need ground-truth masks. The domain class label is available, as we know the centre where the data belong.

\subsection{Meta-train and meta-test objectives}
Our \textit{meta-train} objective contains two components:
\begin{equation}
     \mathcal{L}_{meta-train} = \lambda_{Dice} \mathcal{L}_{Dice}(\mathbf{Y}, \hat{\mathbf{Y}}) + \mathcal{L}_{DT},
\end{equation}
where $\lambda_{Dice}=5$ when labeled data are available.

For the \textit{meta-test} step, the model is expected to: \textbf{1)} accurately predict segmentation masks (by applying the task objective), and \textbf{2)} disentangle $\mathbf{Z}$ and $\mathbf{s}, \mathbf{d}$ to the same level as meta-train sets. A naive strategy for the latter is to use $\mathcal{L}_{DT}$ for meta-test sets. However, as analysed in \cite{antoniou2018train, liu2020shape}, the meta-test step is unstable to train: the gradients from the meta-test loss are second-order statistics of $\psi$ and $\theta$. Our experiments revealed that including the unsupervised losses $\mathcal{L}_{KL}$ and $\mathcal{L}_{HSIC}$ make training even more unstable (even leading to model collapse). In addition, we use one domain for meta-test in experiments, while $\mathcal{L}_{rank}$ requires multiple domains. According to \cite{locatello2019challenging, liu2020metrics}, considering fixed learning and design biases, the level of disentanglement can be proxied by the reconstruction quality (with ground-truth image $\mathbf{X}$) and the domain classification accuracy (with ground-truth label $\mathbf{c}$). Hence, we adopt as the meta-test loss:
\begin{equation}
         \mathcal{L}_{meta-test} =  \lambda_{Dice} \mathcal{L}_{Dice}(\mathbf{Y}, \hat{\mathbf{Y}}) + \lambda_{rec} \mathcal{L}_{rec}{(\mathbf{X}, \hat{\mathbf{X}})} + \lambda_{cls} \mathcal{L}_{cls}{(\mathbf{c}, \hat{\mathbf{c}})}.
\end{equation}
Note that for unlabeled data, $\mathcal{L}_{rec}$ and $\mathcal{L}_{cls}$ do not need ground-truth masks.

\section{Experiments}
\subsection{Tasks and datasets}
\textbf{Multi-centre, multi-vendor \& multi-disease cardiac image segmentation (M$\&$Ms) dataset \cite{mnms}:}
The M$\&$Ms challenge dataset contains 320 subjects. Subjects were scanned at 6 clinical centres in 3 different countries using 4 different magnetic resonance scanner vendors (Siemens, Philips, GE, and Canon) i.e. domains A, B, C and D. For each subject, only the end-systole and end-diastole phases are annotated. Voxel resolutions range from $0.85\times 0.85\times 10$ mm to $1.45\times 1.45\times 9.9$ mm. Domain A contains 95 subjects. Domain B contains 125 subjects. Both domains C and D contain 50 subjects.

\noindent \textbf{Spinal cord gray matter segmentation (SCGM) dataset \cite{prados2017spinal}:}
The data from SCGM \cite{prados2017spinal} are collected from 4 different medical centres with different MRI systems (Philips Achieva, Siemens Trio, Siemens Skyra) i.e. domains 1, 2, 3 and 4. The voxel resolutions range from $0.25\times0.25\times2.5$ mm to $0.5\times0.5\times5$ mm. Each domain has 10 labeled subjects and 10 unlabelled subjects.

\subsection{Baseline models}
\textbf{nnUNet \cite{isensee2021nnu}:} is a self-adapting framework based on 2D and 3D U-Nets \cite{ronneberger2015u} which does not specifically target domain generalisation. Given a labelled training dataset, nnUNet automatically adapts its model design and hyperparameters to obtain optimal performance. In the M$\&$Ms challenge, methods based on nnUNet achieved the top performance \cite{mnms}. \\
\textbf{SDNet+Aug. \cite{liu2020disentangled}:} disentangles the input image to a spatial anatomy and a non-spatial modality factors. Here we use intensity- and resolution- augmented data in a semi-supervised setting. Compared to our method, ``SDNet+Aug." only poses disentanglement to the latent features without meta-learning. \\
\textbf{LDDG \cite{li2020domain}:} is the latest state-of-the-art model for domain-generalised medical image analysis. It also uses a rank loss and when applied in a fully supervised setting, LDDG achieved the best generalisation performance on SCGM. \\
\textbf{SAML \cite{liu2020shape}:} is another gradient-based meta-learning approach. SAML proposed to constraint the compactness and smoothness properties of segmentation masks across meta-train and meta-test sets in a fully supervised setting.

\subsection{Implementation details}
Models are trained using the Adam optimiser \cite{kingma2014adam} with a learning rate of $2e^{-5}$ for 50K iterations using batch size 4. Images are cropped to $224\times 244$ for M\&Ms  and $144\times 144$ for SCGM. $\bm{F}_\psi$ is a 2D UNet \cite{ronneberger2015u} to extract $\mathbf{Z}$ features with 8 channels of same height and width as input image. We follow the designs of SDNet \cite{chartsias2019factorised} for $\bm{E}_S$, $\bm{T}_\theta$ and $\bm{DE}$. $\bm{E}_D$ has the same architecture as $\bm{E}_S$. Both $\mathbf{s}$ and $\mathbf{d}$ have 8 dimensions. $\bm{T}_C$ is a single fully-connected layer. All models are implemented in PyTorch \cite{paszke2019pytorch} and are trained using an NVidia 2080 Ti GPU. In the semi-supervised setting, we use specific percentages of the subjects as labeled data and the rest as unlabeled data. We use Dice (\%) and Hausdorff Distance \cite{dubuisson1994modified} as the evaluation metrics.

\begin{table}[t]
\centering
\caption{Dice (\%) results and the standard deviations on M\&Ms dataset. For ``SDNet+Aug." and our method, the training data contain all the unlabeled data and 2\% or 5\% of labeled data from source domains. The other models are trained by 2\% or 5\%  labeled data only. Bold numbers denote the best performance.}\label{tab1}
\begin{tabular}{|c|c|c|c|c|c|c|c|}
\hline
\multicolumn{2}{|c|}{\textbf{Source}}     & \textbf{Target}           & \textbf{nnUNet}   & \textbf{SDNet+Aug.}  & \textbf{LDDG}    & \textbf{SAML}  & \textbf{Ours} \\ \cline{1-8}
\multirow{ 5}{*}{2\%} & B,C,D  & A                  & $52.87_{ 19}$             & $54.48_{ 18}$     & $59.47_{ 12}$             & $56.31_{ 13}$    & $\mathbf{66.01_{ 12}}$ \\ \cline{2-8}

& A,C,D       & B                                   & $64.63_{ 17}$             & $67.81_{ 14}$     & $56.16_{ 14}$             & $56.32_{ 15}$    & $\mathbf{72.72_{ 10}}$ \\ \cline{2-8}

& A,B,D       & C                                   & $72.97_{ 14}$             & $76.46_{ 12}$     & $68.21_{ 11}$             & $75.70_{ 8.7}$   & $\mathbf{77.54_{ 10}}$  \\ \cline{2-8}

& A,B,C       & D                                   & $73.27_{ 11}$             & $74.35_{ 11}$     & $68.56_{ 10}$             & $69.94_{ 9.8}$   & $\mathbf{75.14_{ 8.4}}$  \\ \cline{2-8}

& \multicolumn{2}{|c|}{\textbf{Average}}            & $65.94_{ 8.3}$             & $68.28_{ 8.6}$     & $63.16_{ 5.4}$             & $64.57_{ 8.5}$    & $\mathbf{72.85_{ 4.3}}$ \\ \cline{2-8}
\hline
\multirow{ 5}{*}{5\%} & B,C,D   & A                 & $65.30_{ 17}$             & $71.21_{ 13}$     & $66.22_{ 9.1}$            & $67.11_{ 10}$    & $\mathbf{72.40_{ 12}}$ \\ \cline{2-8}

& A,C,D       & B                                   & $79.73_{ 10}$             & $77.31_{ 10}$     & $69.49_{ 8.3}$            & $76.35_{ 7.9}$   & $\mathbf{80.30_{ 9.1}}$ \\ \cline{2-8}

& A,B,D       & C                                   & $78.06_{ 11}$             & $81.40_{ 8.0}$    & $73.40_{ 9.8}$            & $77.43_{ 8.3}$   & $\mathbf{82.51_{ 6.6}}$  \\ \cline{2-8}    

& A,B,C       & D                                   & $81.25_{ 8.3}$            & $79.95_{ 7.8}$    & $75.66_{ 8.5}$            & $78.64_{ 5.8}$   & $\mathbf{83.77_{ 5.1}}$ \\ \cline{2-8}

& \multicolumn{2}{|c|}{\textbf{Average}}            & $76.09_{ 6.3}$             & $77.47_{ 3.9}$    & $71.29_{ 3.6}$            & $74.88_{ 4.6}$   & $\mathbf{79.75_{ 4.4}}$  \\ \cline{2-8}
\hline
\end{tabular}
\end{table}

\begin{table}[t]
\centering
\caption{Hausdorff distance results and the standard deviations on M\&Ms dataset. For ``SDNet+Aug." and our method, the training data contain all the unlabeled data and 2\% or 5\% of labeled data from source domains. The other models are trained by 2\% or 5\%  labeled data only. Bold numbers denote the best performance.}\label{tab1_hsd}
\begin{tabular}{|c|c|c|c|c|c|c|c|}
\hline
\multicolumn{2}{|c|}{\textbf{Source}}     & \textbf{Target}           & \textbf{nnUNet}   & \textbf{SDNet+Aug.}  & \textbf{LDDG}    & \textbf{SAML}  & \textbf{Ours} \\ \cline{1-8}
\multirow{ 5}{*}{2\%} & B,C,D  & A                  & $26.48_{ 7.5}$             & $24.69_{ 7.0}$     & $25.56_{ 5.9}$             & $25.57_{ 5.7}$    & $\mathbf{23.55_{ 6.5}}$ \\ \cline{2-8}

& A,C,D       & B                                   & $23.11_{ 6.8}$             & $21.84_{ 6.2}$     & $25.44_{ 5.2}$             & $24.91_{ 5.5}$    & $\mathbf{19.95_{ 6.3}}$ \\ \cline{2-8}

& A,B,D       & C                                   & $16.75_{ 4.6}$             & $16.57_{ 4.2}$     & $18.98_{ 3.9}$             & $16.46_{ 3.5}$   & $\mathbf{16.29_{ 4.0}}$  \\ \cline{2-8}
& A,B,C       & D                                   & $17.51_{ 4.9}$             & $17.57_{ 4.1}$     & $18.08_{ 3.8}$             & $17.94_{ 3.8}$   & $\mathbf{17.48_{ 4.7}}$  \\ \cline{2-8}

& \multicolumn{2}{|c|}{\textbf{Average}}            & $20.96_{ 4.0}$             & $20.17_{ 3.3}$     & $22.02_{ 3.5}$             & $21.22_{ 4.1}$    & $\mathbf{19.32_{ 2.8}}$ \\ \cline{2-8}
\hline
\multirow{ 5}{*}{5\%} & B,C,D   & A                 & $23.04_{ 6.7}$             & $22.84_{ 6.3}$     & $23.35_{ 5.7}$            & $23.10_{ 5.9}$    & $\mathbf{22.55_{ 6.6}}$ \\ \cline{2-8}

& A,C,D       & B                                   & $\mathbf{18.18_{ 4.7}}$             & $20.26_{ 5.5}$     & $20.56_{ 4.7}$            & $18.97_{ 4.9}$   & $19.37_{ 6.4}$ \\ \cline{2-8}

& A,B,D       & C                                   & $16.44_{ 4.2}$             & $16.22_{ 3.9}$    & $17.14_{ 3.3}$            & $16.29_{ 3.2}$   & $\mathbf{15.77_{ 3.8}}$  \\ \cline{2-8}    

& A,B,C       & D                                   & $15.24_{ 4.2}$            & $15.15_{ 3.3}$    & $15.80_{ 3.2}$            & $15.58_{ 3.2}$   & $\mathbf{14.24_{ 2.8}}$ \\ \cline{2-8}

& \multicolumn{2}{|c|}{\textbf{Average}}            & $18.22_{ 3.0}$             & $18.62_{ 3.1}$    & $19.21_{ 3.0}$            & $18.49_{ 2.9}$   & $\mathbf{17.98_{ 3.2}}$  \\ \cline{2-8}
\hline
\end{tabular}
\end{table}

\subsection{Results and discussion}
Tables \ref{tab1}, \ref{tab1_hsd}, \ref{tab2} and \ref{tab2_hsd} show that we consistently achieve the best generalisation performance on cardiac and gray matter segmentation. Particularly in the low data regime we improve Dice by $\approx 5$\% on M\&Ms and $\approx 3$\% on SCGM compared to the best performing baseline. For 100\% annotations in M\&Ms (see Appendix), our model still outperforms the baselines. We also show the qualitative results in Appendix, where the improved performance is visually observed.

\noindent \textbf{M\&Ms}: Compared to ``SDNet+Aug." which can also use (due to disentanglement) unlabeled data, our model performs consistently better. The results agree with the conclusion in \cite{Montero2021genealization}: without specific designs tuned to the tasks, disentanglement can not provide guaranteed generalisation ability. For LDDG and SAML, the generalisation performance significantly drops with small amounts of labeled data. Note that nnUNet adapts the model design per each run/training set. However, adapting the model design for different training data limits the scalability of nnUNet. In the Appendix, we also show that nnUNet possibly overfits the source domains in some cases.

\begin{table}[t]
\centering
\caption{Dice (\%) results and the standard deviations on SCGM dataset. For ``SDNet+Aug." and our method, the training data contain all the unlabeled data and 20\% or 100\% of labeled data from source domains. The other models are trained by 20\% or 100\% of labeled data only. Bold numbers denote the best performance.}\label{tab2}
\begin{tabular}{|c|c|c|c|c|c|c|c|}
\hline
\multicolumn{2}{|c|}{\textbf{Source}}     & \textbf{Target}           & \textbf{nnUNet}               & \textbf{SDNet+Aug.}                & \textbf{LDDG}                & \textbf{SAML}                           & \textbf{Ours} \\ \cline{1-8}
\multirow{ 5}{*}{20\%} & 2,3,4  & 1                 & $59.07_{ 21}$        & $83.07_{ 16}$               & $77.71_{ 9.1}$      & $78.71_{ 25}$                  & $\mathbf{87.45_{ 6.3}}$ \\ \cline{2-8}

& 1,3,4       & 2                                   & $69.94_{ 12}$        & $80.01_{ 5.2}$              & $44.08_{ 12}$       & $75.58_{ 12}$                  & $\mathbf{81.05_{ 5.2}}$ \\ \cline{2-8}

& 1,2,4       & 3                                   & $60.25_{ 7.2}$       & $58.57_{ 10}$               & $48.04_{ 5.5}$      & $54.36_{ 7.6}$                 & $\mathbf{61.85_{ 7.3}}$ \\ \cline{2-8}    

& 1,2,3       & 4                                   & $70.13_{ 4.3}$       & $85.27_{ 2.2}$              & $83.42_{ 2.7}$      & $85.36_{ 2.8}$                 & $\mathbf{87.96_{ 2.1}}$ \\ \cline{2-8}

& \multicolumn{2}{|c|}{\textbf{Average}}            & $64.85_{ 5.2}$    & $76.73_{ 11}$           & $63.31_{ 17}$   & $73.50_{ 12}$               & $\mathbf{79.58_{ 11}}$ \\ \cline{2-8}
\hline
\multirow{ 5}{*}{100\%} & 2,3,4  & 1                & $75.27_{ 8.3}$    & $\mathbf{90.25_{ 4.5}}$  & $88.21_{ 4.9}$   & $90.22_{ 5.6}$              & $90.01_{ 4.9}$ \\ \cline{2-8}

& 1,3,4       & 2                                   & $76.32_{ 2.9}$    & $84.13_{ 4.2}$           & $83.76_{ 3.1}$   & $\mathbf{86.65_{ 3.5}}$     & $85.48_{ 2.3}$ \\ \cline{2-8}

& 1,2,4       & 3                                   & $62.59_{ 6.9}$    & $62.18_{ 10}$            & $56.11_{ 9.3}$   & $58.27_{ 9.4}$              & $\mathbf{64.23_{ 9.7}}$ \\ \cline{2-8}

& 1,2,3       & 4                                   & $71.87_{ 2.5}$    & $88.93_{ 1.9}$           & $89.08_{ 2.7}$   & $88.66_{ 2.6}$              & $\mathbf{89.26_{ 2.5}}$ \\ \cline{2-8}

& \multicolumn{2}{|c|}{\textbf{Average}}            & $71.51_{ 5.4}$    & $81.37_{ 11}$           & $79.29_{ 13}$   & $80.95_{ 13}$              & $\mathbf{82.25_{ 11}}$ \\ \cline{2-8}
\hline
\end{tabular}
\end{table}

\begin{table}[t]
\centering
\caption{Hausdorff distance results and the standard deviations on SCGM dataset. For ``SDNet+Aug." and our method, the training data contain all the unlabeled data and 20\% or 100\% of labeled data from source domains. The other models are trained by 20\% or 100\% of labeled data only. Bold numbers denote the best performance.}\label{tab2_hsd}
\begin{tabular}{|c|c|c|c|c|c|c|c|}
\hline
\multicolumn{2}{|c|}{\textbf{Source}}     & \textbf{Target}           & \textbf{nnUNet}               & \textbf{SDNet+Aug.}                & \textbf{LDDG}                & \textbf{SAML}                           & \textbf{Ours} \\ \cline{1-8}
\multirow{ 5}{*}{20\%} & 2,3,4  & 1                 & $3.09_{ 0.25}$        & $1.52_{ 0.33}$               & $1.75_{ 0.26}$      & $1.53_{ 0.38}$                  & $\mathbf{1.50_{ 0.30}}$ \\ \cline{2-8}

& 1,3,4       & 2                                   & $3.16_{ 0.09}$        & $1.97_{ 0.16}$              & $2.73_{ 0.33}$       & $2.07_{ 0.35}$                  & $\mathbf{1.91_{ 0.16}}$ \\ \cline{2-8}

& 1,2,4       & 3                                   & $3.38_{ 0.27}$       & $2.45_{ 0.27}$               & $2.67_{ 0.25}$      & $2.52_{ 0.24}$                 & $\mathbf{2.23_{ 0.23}}$ \\ \cline{2-8}    

& 1,2,3       & 4                                   & $4.31_{ 0.14}$       & $2.34_{ 0.21}$              & $2.37_{ 0.14}$      & $2.30_{ 0.18}$                 & $\mathbf{2.22_{ 0.13}}$ \\ \cline{2-8}

& \multicolumn{2}{|c|}{\textbf{Average}}            & $3.49_{ 0.49}$    & $2.07_{ 0.36}$           & $2.38_{ 0.39}$   & $2.11_{ 0.37}$               & $\mathbf{1.97_{ 0.30}}$ \\ \cline{2-8}
\hline
\multirow{ 5}{*}{100\%} & 2,3,4  & 1                & $3.26_{ 0.21}$    & $\mathbf{1.37_{ 0.25}}$  & $1.50_{ 0.23}$   & $1.43_{ 0.36}$              & $1.43_{ 0.29}$ \\ \cline{2-8}

& 1,3,4       & 2                                   & $3.19_{ 0.09}$    & $1.88_{ 0.16}$           & $2.19_{ 0.19}$   & $\mathbf{1.80_{ 0.19}}$     & $1.81_{ 0.15}$ \\ \cline{2-8}

& 1,2,4       & 3                                   & $3.37_{ 0.27}$    & $2.34_{ 0.24}$            & $2.64_{ 0.28}$   & $2.43_{ 0.33}$              & $\mathbf{2.23_{ 0.32}}$ \\ \cline{2-8}

& 1,2,3       & 4                                   & $4.30_{ 0.15}$    & $2.13_{ 0.17}$           & $2.12_{ 0.15}$   & $2.15_{ 0.15}$              & $\mathbf{2.11_{ 0.13}}$ \\ \cline{2-8}

& \multicolumn{2}{|c|}{\textbf{Average}}            & $3.53_{ 0.45}$    & $1.93_{ 0.36}$           & $2.11_{ 0.41}$   & $1.95_{ 0.38}$              & $\mathbf{1.92_{ 0.31}}$ \\ \cline{2-8}
\hline
\end{tabular}
\end{table}

\noindent \textbf{SCGM}: We obtain consistent improvements also on SCGM, demonstrating application in other organs. Our model benefits from the additional 10 unlabeled subjects of each domain leading to better performance overall.

\subsection{Ablation study}
Here we conduct ablations on key losses crucial to disentanglement and the extraction of good anatomical features for good generalisation performance. We omit ablations on the KL losses as \cite{higgins2016beta, ilse2020diva} showcase that variational encoding helps to learn robust vector representation for better generalisation. To illustrate that $\mathcal{L}_{rank}$ helps to disentangle $\mathbf{Z}$ to $(\mathbf{s},\mathbf{d})$, and improves performance, we use Distance Correlation ($DC$) \cite{liu2020metrics} to measure disentanglement (lower $DC$ means a higher level of disentanglement). For M\&Ms 5\% cases, without $\mathcal{L}_{rank}$, the average $DC$ on the test dataset between $\mathbf{Z}$ and $(\mathbf{s},\mathbf{d})$ is 0.22 (an increase compared to 0.19 with $\mathcal{L}_{rank}$), and the average Dice is 78.54\% (a decrease compared to 79.75\% with $\mathcal{L}_{rank}$).
We also ablate $\mathcal{L}_{cls}$ and $\mathcal{L}_{HSIC}$. The proposed model on M\&Ms 5\% cases had an average Dice 79.75\% but without $\mathcal{L}_{cls}$, average Dice drops to 77.45\% and without $\mathcal{L}_{HSIC}$, average Dice drops to 77.86\%. 

\section{Conclusion}
We have presented a novel semi-supervised meta-learning framework for domain generalisation. Using disentanglement our approach models domain shifts, and thanks to our reconstruction approach to disentanglement, our model can be trained also with unlabeled data. By applying the designed constraints (including the low-rank regularisation) to the gradient-based meta-learning approach, the model extracts robust anatomical features useful for predicting segmentation masks in a semi-supervised manner. Extensive quantitative results, especially when insufficient annotated data are available, indicate remarkable improvements compared to previous state-of-the-art approaches. The performance of our method might improve with the use of additional unlabeled data from other domains, which we leave as future work. 

\section{Acknowledgement}
This work was supported by the University of Edinburgh, the Royal Academy of Engineering and Canon Medical Research Europe by a PhD studentship to Xiao Liu. This work was partially supported by the Alan Turing Institute under the EPSRC grant EP/N510129/1. We thank Nvidia for donating a Titan-X GPU.  S.A. Tsaftaris acknowledges the support of Canon Medical and the Royal Academy of Engineering and the Research Chairs and Senior Research Fellowships scheme (grant RCSRF1819\textbackslash8\textbackslash25).

%
\bibliographystyle{splncs04}
\bibliography{references}

\newpage

\section{Appendix}
\subsection{More quantitative results}
In Table \ref{tabA1} and Table \ref{tab1_hsd}, we report the Dice (\%) and Hausdorff Distance results on the cases of giving 100\% labeled data for each model. For M\&Ms, the unlabeled data (of labeled subjects) from phases between end-systole and end-diastole phases still gives the proposed model slightly better performance i.e. 0.65\% of improvement compared with the best baseline model.

\begin{table}[ht]
\centering
\renewcommand{\thetable}{A\arabic{table}}
\caption{Dice (\%) results and the standard deviations on M\&Ms dataset. For ``SDNet+Aug." and our method, the training data contains all the unlabeled data and all labeled data from source domains. The other models are trained by labeled data only. Bold numbers denote the best performance.}\label{tabA1}
\begin{tabular}{|c|c|c|c|c|c|c|c|}
\hline
\multicolumn{2}{|c|}{\textbf{Source}}     & \textbf{Target}           & \textbf{nnUNet}               & \textbf{SDNet+Aug.}                & \textbf{LDDG}                & \textbf{SAML}                           & \textbf{Ours} \\ \cline{1-8}

\multirow{ 5}{*}{100\%} & B,C,D  & A                & $80.84_{\pm 11}$             & $81.50_{\pm 7.7}$    & $82.62_{\pm 6.3}$            & $81.33_{\pm 7.2}$   & $\mathbf{83.21_{\pm 7.4}}$ \\ \cline{2-8}

& A,C,D       & B                                   & $\mathbf{86.76_{\pm 5.8}}$   & $85.04_{\pm 6.1}$    & $85.68_{\pm 5.7}$            & $84.15_{\pm 5.9}$   & $86.53_{\pm 5.3}$ \\ \cline{2-8}

& A,B,D       & C                                   & $84.92_{\pm 7.1}$            & $85.64_{\pm 6.5}$    & $86.49_{\pm 6.3}$            & $84.52_{\pm 6.2}$   & $\mathbf{87.22_{\pm 6.1}}$  \\ \cline{2-8}

& A,B,C       & D                                   & $86.94_{\pm 5.9}$            & $84.96_{\pm 5.2}$    & $86.73_{\pm 6.1}$            & $83.96_{\pm 5.9}$   & $\mathbf{87.16_{\pm 4.9}}$  \\ \cline{2-8}
&  \multicolumn{2}{|c|}{\textbf{Average}}           & $84.87_{\pm 2.5}$            & $84.29_{\pm 1.6}$    & $85.38_{\pm 1.6}$            & $83.49_{\pm 1.3}$   & $\mathbf{86.03_{\pm 1.7}}$  \\
\hline
\end{tabular}
\end{table}

\begin{table}[ht]
\centering
\renewcommand{\thetable}{A\arabic{table}}
\caption{Hausdorff distance results and the standard deviations on M\&Ms dataset. For ``SDNet+Aug." and our method, the training data contains all the unlabeled data and all labeled data from source domains. The other models are trained by labeled data only. Bold numbers denote the best performance.}\label{tabA1_hsd}
\begin{tabular}{|c|c|c|c|c|c|c|c|}
\hline
\multicolumn{2}{|c|}{\textbf{Source}}     & \textbf{Target}           & \textbf{nnUNet}               & \textbf{SDNet+Aug.}                & \textbf{LDDG}                & \textbf{SAML}                           & \textbf{Ours} \\ \cline{1-8}

\multirow{ 5}{*}{100\%} & B,C,D  & A                & $17.86_{\pm 5.5}$             & $17.39_{\pm 4.5}$    & $17.48_{\pm 4.1}$            & $17.70_{\pm 4.2}$   & $\mathbf{17.28_{\pm 3.9}}$ \\ \cline{2-8}

& A,C,D       & B                                   & $\mathbf{14.82_{\pm 3.4}}$   & $15.55_{\pm 3.7}$    & $15.42_{\pm 3.4}$            & $16.05_{\pm 3.7}$   & $14.99_{\pm 3.6}$ \\ \cline{2-8}

& A,B,D       & C                                   & $13.72_{\pm 3.3}$            & $13.67_{\pm 3.0}$    & $13.52_{\pm 2.8}$            & $14.21_{\pm 3.3}$   & $\mathbf{13.11_{\pm 2.8}}$  \\ \cline{2-8}

& A,B,C       & D                                   & $12.81_{\pm 3.4}$            & $13.64_{\pm 2.9}$    & $13.11_{\pm 3.0}$            & $14.12_{\pm 2.8}$   & $\mathbf{12.72_{\pm 2.6}}$  \\ \cline{2-8}
&  \multicolumn{2}{|c|}{\textbf{Average}}           & $14.80_{\pm 1.9}$            & $15.06_{\pm 1.6}$    & $14.88_{\pm 1.7}$            & $15.52_{\pm 1.5}$   & $\mathbf{14.53_{\pm 1.8}}$  \\
\hline
\end{tabular}
\end{table}

\subsection{Disentanglement networks}
We use AdaIN module in the decoder $\bm{DE}$ to combine $\mathbf{Z}$ and $\mathbf{s},\mathbf{d}$. Each AdaIN layer performs the following operation: 

\begin{equation}
    \mathrm{AdaIN} = \sigma(\mathbf{s}_{i}, \mathbf{d}_{i})\frac{\mathbf{Z}_{i} - \mu(\mathbf{Z}_{i})}{\sigma(\mathbf{Z}_{i})} + \mu(\mathbf{s}_{i}, \mathbf{d}_{i}),
\end{equation}
where each feature map $\mathbf{Z}_{i}$ is first normalized separately, and then is scaled and shifted based on the style representation and domain representation $\mathbf{s}_{i}, \mathbf{d}_{i}$ mean and standard deviation scalars.

\subsection{Analysis of M\&Ms data}
To explore why nnUNet can outperform other models on cases of A,C,D to B and A,B,C to D, we train nnUNet models with 100\% labeled data only from domain B or domain D. Then we test the models with data from domain A, B, C and D. As shown in Table \ref{tabmnms}, the model trained on domain B can achieve 85.81\% Dice on domain D (highest Dice compared to A and C). Also, the model trained on domain D can achieve 84.35\% Dice on domain B (highest Dice compared to A and C). Hence, nnUNet possibly overfits the source domains e.g. B, to achieve the best performance on domains (e.g. D) similar to B. However, other methods have to generalise to distinct domains i.e. A and C, which causes slightly worse performance on the domains similar to source domains.

\begin{table}[t]
\centering
\renewcommand{\thetable}{A\arabic{table}}
\caption{Dice (\%) results and the standard deviations on M\&Ms dataset.}\label{tabmnms}
\begin{tabular}{|c|c|c|c|c|c|}
\hline
\multicolumn{2}{|c|}{\textbf{Source}}     & \textbf{Target A}          & \textbf{Target B}         & \textbf{Target C}         & \textbf{Target D}   \\ \cline{1-6}

\multirow{ 2}{*}{100\%} & Trained on B      & $78.82_{11}$      & $94.58_{4.3}$    & $83.60_{8.0}$    & $85.81_{6.8}$   \\\cline{2-6}

                        & Trained on D      & $80.03_{10}$      & $84.35_{7.0}$    & $84.01_{8.9}$    & $94.74_{5.4}$   \\\cline{2-6}
\hline
\end{tabular}
\end{table}

\subsection{Qualitative results}
\begin{figure}[t]
\renewcommand{\thefigure}{A\arabic{figure}}
\includegraphics[width=\textwidth]{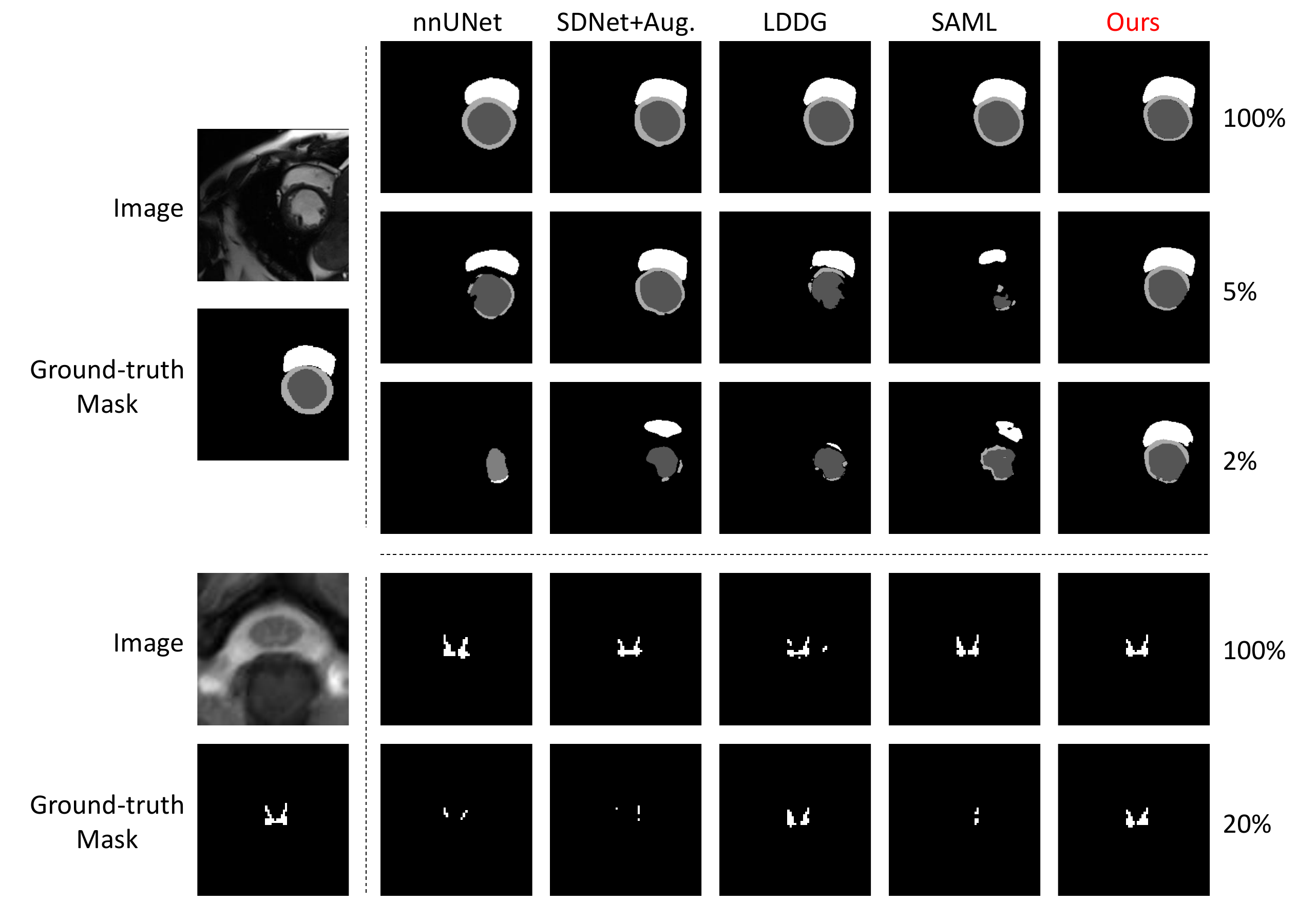}
\centering
\caption{We show the example images and predicted segmentation masks of each model for different cases. When training the baseline models with less labeled data, the performance drops significantly. In contrast, our model can produce satisfactory masks in every case.} \label{fig::example}
\end{figure}

We show the qualitative results, i.e. predicted segmentation masks, in Fig. \ref{fig::example}.

\end{document}